\documentclass[letterpaper]{article} 
\usepackage{aaai2027}  
\usepackage[hyphens]{url}  
\usepackage{graphicx} 
\usepackage{natbib}  
\usepackage{caption} 
\usepackage{algorithm}
\usepackage{algorithmic}
\usepackage{amsmath}
\usepackage{amssymb}
\usepackage{placeins}
\usepackage{multirow}

\usepackage{newfloat}
\usepackage{listings}
\DeclareCaptionStyle{ruled}{labelfont=normalfont,labelsep=colon,strut=off} 
\floatstyle{ruled}
\newfloat{listing}{tb}{lst}{}
\floatname{listing}{Listing}

\usepackage{booktabs}

\usepackage{url}
\usepackage{booktabs}
\usepackage{multirow}
\usepackage{makecell}
\usepackage{graphicx}
\usepackage[table]{xcolor}

\graphicspath{{Figures/}}

\definecolor{gnncolor}{RGB}{232,241,252}
\definecolor{dfcolor}{RGB}{255,244,204}
\definecolor{rfmcolor}{RGB}{255,230,230}

\title{Towards Unified Multimodal Graph Foundation Model:\\ A Bridge-Router-Adapter Based Approach}
\author{
    Sirui Zhang,
    Yubing Zhou,
    Xunkai Li,
    Zekai Chen,
    Shumeng Li,
    Wang Luo,
    Yinlin Zhu,
    Yujin Gao,
    Rong-Hua Li\corresponding
}
\affiliations{
    Beijing Institute of Technology\\
    Beijing, China\\
}

\begin{document}

\maketitle

\begin{abstract}
Multimodal graphs couple node attributes in different modalities, such as text and images, with relational structure, enabling topological structure and cross-modality attributes to be modeled jointly. Multimodal graph foundation models seek unified representations from such data that transfer across different graph domains and downstream tasks. However, existing methods exhibit two fundamental limitations.
\textbf{(1) Cross-Scope Context Entanglement.}
They merge scope-specific graph contexts into a unified representation, obscuring their distinctions during multimodal construction.
\textbf{(2) Scope-Ignorant Modality Routing.}
They route modalities within a fixed graph scope, overlooking how modality relevance varies across neighborhood ranges. To address these challenges, we propose \textbf{\textsc{BRAIN}}, a unified model that focuses on graph context that combines neighborhood scope with modality composition. BRAIN comprises a scope-conditioned \emph{Bridge} that combines structural information spanning local-to-global neighborhood scopes with different modality compositions; a hierarchical \emph{Router} that estimates the relevance between the scope and the task, and selects compositions separately within each scope, allowing modality utility to vary with graph range; and a lightweight residual \emph{Adapter} that further specializes the routed embedding for downstream prediction. \textsc{BRAIN} is trained through multi-graph pretraining followed by task-specific adaptation. Experiments across nine datasets and four task families demonstrate its broad effectiveness, improving node-classification and link-prediction performance by up to \textbf{4.73\%} relative to the strongest baseline, while achieving an average relative improvement of \textbf{14.72\%} across four graph-to-text and two graph-to-image metrics.
\end{abstract}


\begin{figure*}[t]
    \centering
    \includegraphics[width=\textwidth]{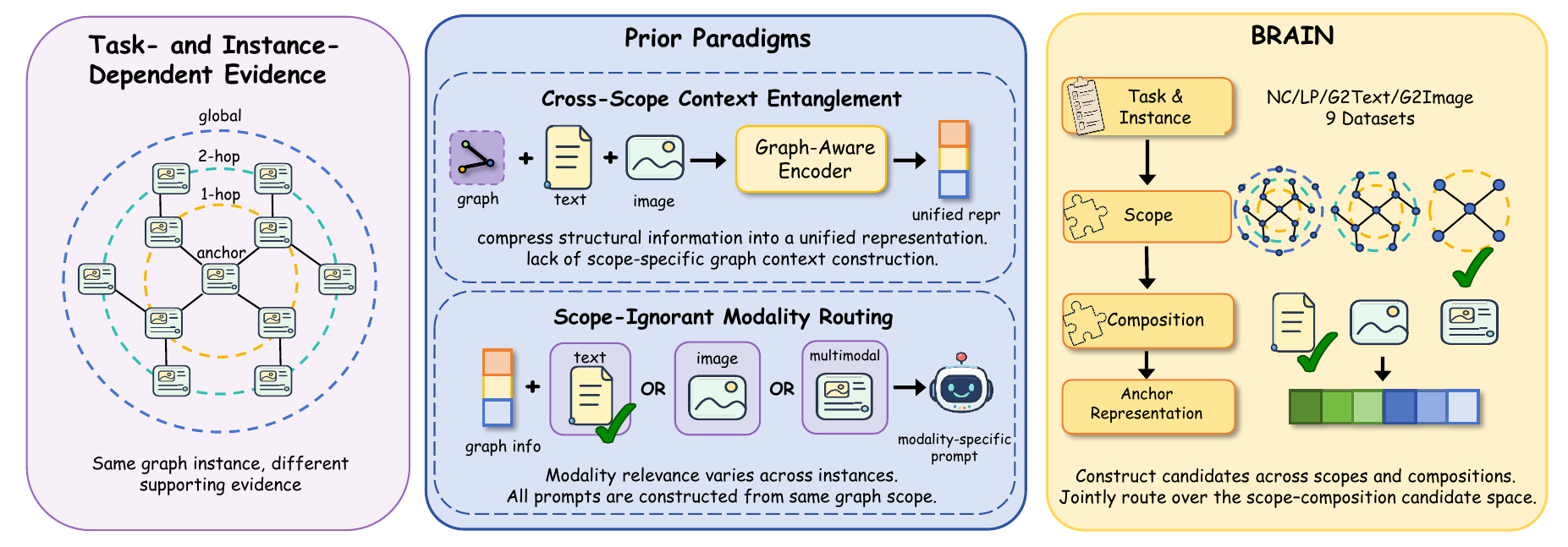}
    \caption{Comparison of context construction paradigms. Prior methods
    either construct a unified representation or adapt multimodal
    processing under a prescribed structural organization. \textsc{BRAIN}
    constructs and routes a candidate pool across neighborhood scopes
    and modality compositions according to the task and instance.}
    \label{fig:motivation}
\end{figure*}

\section{Introduction}
\label{sec:introduction}

Multimodal graphs (MMGs) represent entities as nodes, relations as edges, and text and image attributes as node content. Their relational structure supports recommendation, graph-aware understanding, and graph-grounded generation \cite{wei2019mmgcn,ning2025graph4mm,feng2025gwm,wan2026openmag}. Unlike isolated samples, each node is interpreted through both its attributes and related context.

Graph foundation models seek representations transferable across domains \cite{mao2024graphfoundation,liu2024ofa}. Recent methods learn unified MMG embeddings, inject graph neighborhoods into language models, or integrate graph encoders with language and vision--language models \cite{he2025unigraph2,yoon2023mmgl,chen2024llaga,fan2025mlaga,fang2025graphgpto}. However, the appropriate neighborhood scope and modality composition vary across tasks and instances.

As summarized in Fig.~\ref{fig:motivation}, existing methods exhibit two fundamental limitations. \textbf{(1) Cross-Scope Context Entanglement.}  Some methods collapse scope-specific graph context into the unified representation, instead of explicitly preserving and distinguishing it throughout multimodal representation construction. UniGraph2 aligns propagated modality features in a unified space, while Graph4MM and PLANET incorporate multi-hop structure through cross-modal interaction \cite{he2025unigraph2,ning2025graph4mm,liu2026planet}. As different neighborhood ranges are absorbed into one output, local, broader, and compressed global context cannot remain distinguishable, limiting scope-specific representation construction.

\textbf{(2) Scope-ignorant Modality Routing.} Many current methods adapt modality usage under a fixed graph scope, failing to capture how modality relevance changes across local-to-global neighborhoods. Mario routes among modality instruction views, while DiP adjusts intra-modal propagation and inter-modal aggregation \cite{sun2026mario,hong2026dip}. However, their structural range remains prescribed rather than jointly selected with modality composition. They determine which modalities to use, but not the graph range in which to interpret them, although modality utility may vary across neighborhood ranges.

To address these challenges, we propose \textbf{\textsc{BRAIN}}, a \textbf{B}ridge--\textbf{R}outer--\textbf{A}dapter \textbf{I}nfusion \textbf{N}etwork centered on graph context. \textsc{BRAIN} defines context as $c=(s,m)$, where $s$ denotes neighborhood scope and $m$ denotes modality composition, and models their task- and instance-dependent relevance. Its central intuition is to separate context construction, conditional selection, and task specialization: it constructs candidates across local-to-global scopes and modality compositions, routes them according to the task and instance, and specializes the routed representation. Multi-graph pretraining learns transferable scope representations, followed by task-specific adaptation.

The scope-conditioned \emph{Bridge} constructs a graph-aware representation for each neighborhood scope and derives text-dominant, image-dominant, and multimodal candidates from it. This preserves scope identity and separates scope aggregation from composition specialization, enabling multiple modality interpretations without repeatedly reconstructing structural context.

The candidate pool is processed by a hierarchical \emph{Router}, which first estimates scope relevance and then selects modality compositions separately within each scope. This explicitly conditions modality preference on graph range. A lightweight residual \emph{Adapter} subsequently specializes the routed representation while retaining transferable information. These modules jointly preserve context during local-to-global construction, coupling it with scope-dependent modality routing.

\textit{Our contributions.} (1) \textit{New Perspective.}We formulate \emph{graph-conditioned context relevance} as task- and instance-dependent modeling of neighborhood scope and modality composition. (2) \textit{New Method.} We introduce \textsc{BRAIN}, combining scope-conditioned candidate construction, hierarchical scope-first routing, and residual adaptation. (3) \textit{SOTA Performance.} Across nine datasets and four task families, \textsc{BRAIN} ranks first on nine of ten discriminative comparisons and achieves an average relative improvement of \textbf{14.72\%} across four graph-to-text and two graph-to-image metrics.

\begin{figure*}[t]
\centering
\includegraphics[width=\textwidth]{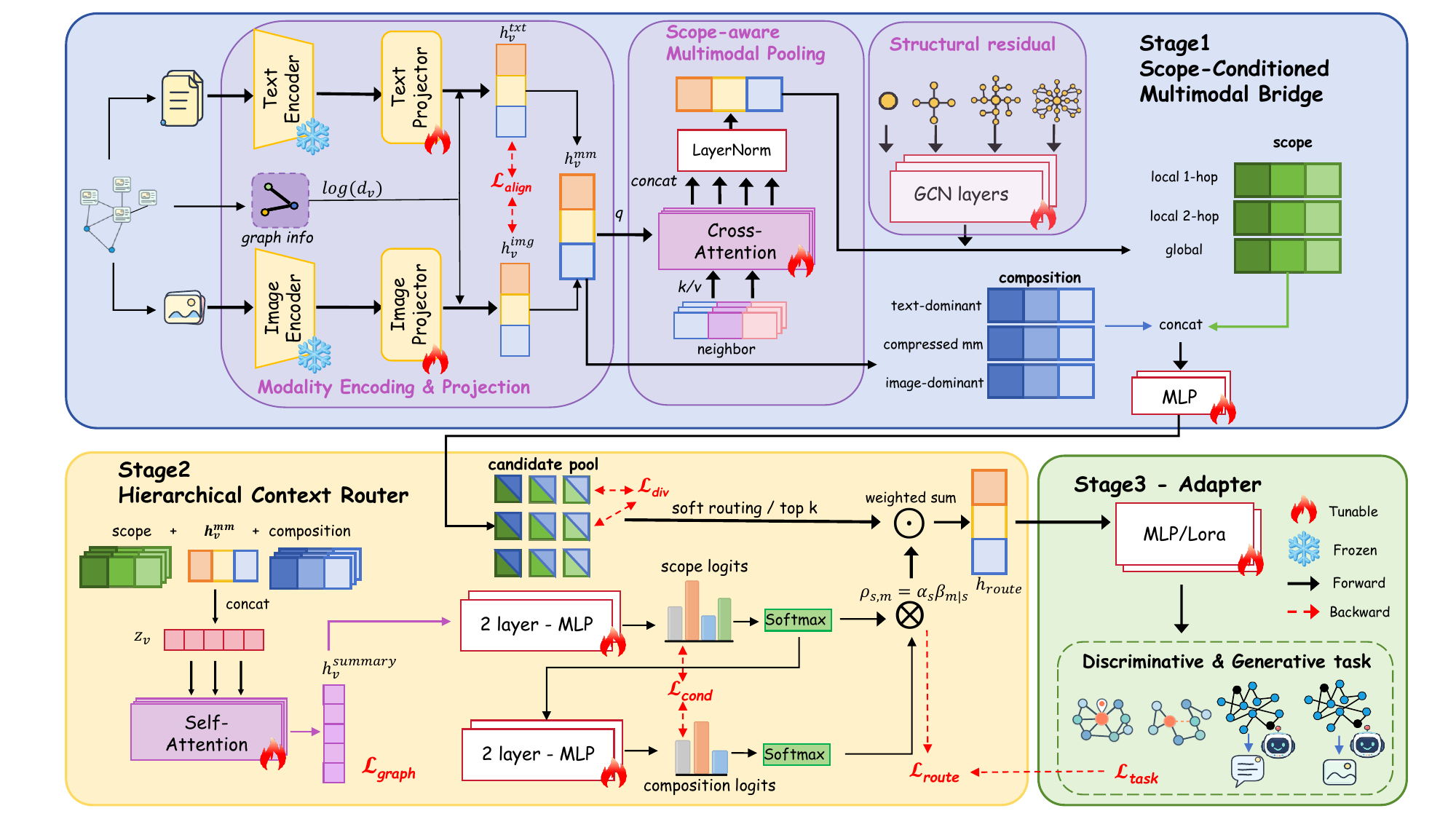}
\caption{\textsc{BRAIN} uses two training stages: Stage~1 pretrains reusable scope representations, and Stage~2 learns routing, post-routing adaptation, and task prediction. The Bridge forms a scope--composition candidate pool; the conditional Router estimates $\alpha_s\beta_{m\mid s}$; and a residual Adapter specializes the routed representation before task prediction.}
\label{fig:framework}
\end{figure*}

\section{Preliminaries}
\label{sec:preliminary}

\paragraph{Problem formulation.}
Consider a multimodal attributed graph $G=(\mathcal V,\mathcal E,\mathcal X)$, where $\mathcal X=\{(x_v^{\mathrm{txt}},x_v^{\mathrm{img}})\}_{v\in\mathcal V}$ contains textual and visual node attributes. \textsc{BRAIN} first constructs reusable node-level alternatives and allows a downstream task to read one or more node representations. For an anchor node $v$, let
$\mathcal S=\{\text{1-hop},\text{2-hop},\text{global}\}$ denote the available neighborhood scopes and let $\mathcal M=\{\text{text-dominant},\text{image-dominant},\text{multimodal}\}$ denote the modality compositions. Each context candidate is indexed by
\begin{equation}
c=(s,m)\in\mathcal S\times\mathcal M.
\end{equation}
For task $\tau$, \textsc{BRAIN} models the joint relevance of scope and composition through
\begin{equation}
p(c\mid v,\tau,G)=p(s\mid v,\tau,G)p(m\mid s,v,\tau,G).
\end{equation}
Downstream tasks consume one or more routed node representations through lightweight task-specific heads; tensor organization and task interfaces are detailed in Supplementary Secs.~C.1 and C.5.

\paragraph{Related Work.}
\label{sec:related_work}
We review multimodal graph learning and graph foundation models, which respectively integrate heterogeneous modalities with graph structure and learn transferable representations across graphs and tasks.

\textit{Multimodal graph learning.}
Early methods propagate information over interaction graphs and fuse modality representations \cite{wei2019mmgcn}. Later studies construct modality-aware graphs from visual and textual similarities or improve them through graph freezing and denoising \cite{zhang2021lattice,zhou2023freedom}. Recent work explores heterogeneous neighbor aggregation, cross-modal modeling, and graph-conditioned language models \cite{yoon2023mmgl,ning2025graph4mm,fan2025mlaga,fang2025graphgpto}. Nevertheless, these methods mainly encode a predefined graph view into a single multimodal representation.

\textit{Graph foundation models.}
Graph foundation models learn reusable representations across graphs and tasks \cite{mao2024graphfoundation}. Existing approaches use prompting, structure-aware tokenization, shared encoders, cross-domain alignment, domain adaptation, and topology-aware pretraining \cite{liu2024ofa,chen2024llaga,he2025unigraph2,feng2025gwm,wang2025mdgfm,yuan2025bridge}. Multimodal variants further align graph structure, text, and images within transferable backbones \cite{he2025unigraph2,liu2026planet}. However, most rely on a predetermined structural range rather than selecting among graph scopes and modality compositions.

\section{Methodology}
\label{sec:methods}

\subsection{Overview}
\label{sec:method_overview}

As discussed above, existing methods do not jointly model which graph range is relevant and which modality composition is useful within that range. \textsc{BRAIN} addresses this gap by separating scope-aware candidate construction from task-aware candidate selection.

As illustrated in Fig.~\ref{fig:framework}, \textsc{BRAIN} follows a Bridge--Router--Adapter architecture. Given an anchor node $v$, the Scope-Conditioned Multimodal Bridge first constructs a shared representation $S_{v,s}$ for each graph scope $s \in \mathcal{S}$ by combining anchor-relevant semantic information with connectivity-aware structural information. It then derives text-dominant, image-dominant, and multimodal candidate representations $\{V_{v,s,m}\}_{m \in \mathcal{M}}$ from each shared scope representation. This separation performs contextual aggregation once per scope while retaining distinct alternatives across graph ranges and modality compositions.

For downstream task $\tau$, the Hierarchical Context Router reads the candidate pool in a scope-first manner. It first estimates scope relevance and then predicts a composition distribution conditioned on each scope. Their product aggregates the candidates into a routed representation, which is subsequently specialized by a lightweight task-conditioned Adapter for task heads. \textsc{BRAIN} therefore constructs reusable node-level alternatives and lets each task determine which graph range and modality organization to emphasize.

\textsc{BRAIN} is optimized in two stages. Stage~1 jointly samples anchor nodes from multiple graph datasets and pretrains the shared node- and scope-representation pathway. A symmetric InfoNCE loss aligns the refined textual and visual embeddings of the same anchor, while a graph contrastive loss encourages its graph summary to preserve information consistent with the observed neighborhood scopes. The Stage~1 objective is
\begin{equation}
\label{eq:stage1_objective}
\mathcal{L}_{\mathrm{S1}} =
\mathcal{L}_{\mathrm{align}}+\lambda_G\mathcal{L}_{\mathrm{graph}}.
\end{equation}
This stage establishes reusable multimodal and structural representations before task-dependent routing is introduced.

Stage~2 initializes the shared representation pathway from Stage~1 and jointly optimizes composition specialization, the Router, the post-routing Adapter, and task heads. Candidate-level task losses are converted by a stop-gradient teacher into a target utility distribution that assigns greater probability to candidates with lower loss. The routing loss $\mathcal{L}_{\mathrm{route}}$ aligns the predicted joint distribution with this teacher. The conditional loss $\mathcal{L}_{\mathrm{cond}}$ emphasizes samples whose preferred modality compositions vary across scopes, while $\mathcal{L}_{\mathrm{div}}$ prevents different compositions within the same scope from collapsing to redundant representations. The Stage~2 objective is
\begin{equation}
\label{eq:stage2_objective}
\mathcal{L}_{\mathrm{S2}} =
\mathcal{L}_{\mathrm{task}}
+\lambda_{\mathrm{route}}\mathcal{L}_{\mathrm{route}}
+\lambda_{\mathrm{cond}}\mathcal{L}_{\mathrm{cond}}
+\lambda_{\mathrm{div}}\mathcal{L}_{\mathrm{div}}.
\end{equation}
The two stages thus separate reusable candidate-space learning from task-supervised routing and adaptation. Complete loss definitions and training pseudocode are provided in Supplementary Sec.~D.

\subsection{Scope-Conditioned Multimodal Bridge}
\label{sec:bridge}

The Bridge separates scope aggregation from composition specialization. It first constructs one shared multimodal representation for each scope and then derives text-dominant, image-dominant, and multimodal candidates from that representation. Consequently, contextual aggregation is performed once per scope rather than repeated for every composition.

\paragraph{Scope representation.}
For each anchor--scope pair, the semantic branch selects contextual information conditioned on both the refined anchor embedding and the target graph scope. With scope embedding $e_s$, the scope-conditioned query and semantic readout are defined as
\begin{equation}
\label{eq:semantic_scope_readout}
\begin{aligned}
q_{v,s} &= W_q [\widetilde{h}_v^{\mathrm{mm}} \mathbin{|} e_s], \\
(S_{v,s}^{\mathrm{att}}, \Omega_{v,s}) &= \operatorname{MHA}(q_{v,s}, H_{v,s}^{\mathrm{mm}}, H_{v,s}^{\mathrm{mm}}).
\end{aligned}
\end{equation}
Here, $\Omega_{v,s}$ denotes the attention map. The semantic branch identifies context nodes whose multimodal information is most relevant to anchor $v$ under scope $s$.

In parallel, the structural branch, enabled by default in all reported BRAIN
configurations, propagates information over the induced scoped subgraph and
pools the resulting node representations:
\begin{equation}
\label{eq:structural_scope_readout}
S_{v,s}^{\mathrm{gcn}} = \operatorname{Pool} \left( \operatorname{GCN}(H_{v,s}^{\mathrm{mm}}, A_{v,s}) \right).
\end{equation}
Whereas the semantic branch prioritizes anchor-relevant information, the structural branch preserves connectivity-aware interactions among nodes within the same graph range.

The outputs of the two branches are combined to construct the shared representation of scope $s$:
\begin{equation}
\label{eq:scope_fusion}
S_{v,s} = S_{v,s}^{\mathrm{att}} + \eta_s S_{v,s}^{\mathrm{gcn}},
\end{equation}
where $\eta_s \in (0,1)$ controls the structural contribution. The resulting $S_{v,s}$ provides the shared graph-aware basis from which all modality compositions at scope $s$ are derived.

\paragraph{Scope-aligned candidates.}
Let $\bar{\Omega}_{v,s}$ denote the attention map averaged across heads. \textsc{BRAIN} reuses this node-selection distribution to obtain textual and visual readouts $o_{v,s}^{\mathrm{txt}}$ and $o_{v,s}^{\mathrm{img}}$ from the same context nodes. Each scope--composition candidate is then constructed by
\begin{equation}
\label{eq:candidate_construction}
V_{v,s,m} = \operatorname{Mod} \left( S_{v,s}, \psi_m(\widetilde{h}_v^{\mathrm{txt}}, \widetilde{h}_v^{\mathrm{img}}, o_{v,s}^{\mathrm{txt}}, o_{v,s}^{\mathrm{img}}), c_m \right).
\end{equation}
Reusing the attention map preserves node-level correspondence between textual and visual information and avoids repeated graph aggregation. The composition function $\psi_m$ organizes anchor and contextual signals according to composition $m$. The text- and image-dominant branches place greater emphasis on their respective modalities, whereas the multimodal branch additionally models cross-modal agreement and difference. Here, $\operatorname{Mod}$ is a bounded residual feature-wise modulation that maps the composition signal and embedding $c_m$ to scale and shift terms applied to $S_{v,s}$. This preserves the shared graph-aware scope representation while enabling composition-specific specialization. All candidates remain modality-complete; ``dominant'' indicates only the principal conditioning modality.

Together, the candidates $\{V_{v,s,m}\}_{s \in \mathcal{S}, m \in \mathcal{M}}$ form the scope--composition candidate pool. The Bridge additionally constructs a graph summary $g_v$ by applying attention pooling to $\{S_{v,s}\}_{s \in \mathcal{S}}$ together with the refined anchor embedding. The graph summary conditions the Router, while the candidate pool provides the alternative node representations available for task-dependent selection. Projection, scoped-context construction, and the full modulation equations are detailed in Supplementary Secs.~C.2--C.3.

\subsection{Hierarchical Context Router and Adapter}
\label{sec:context_router}

Unlike flat or scope-independent routing, \textsc{BRAIN}'s scope-first factorization allows modality preference to vary across graph ranges: it first estimates a scope distribution and then predicts a separate composition distribution conditioned on each scope.

For task $\tau$ with embedding $e_\tau$, the Router first constructs a task-aware query:
\begin{equation}
\label{eq:routing_query}
r_v = \operatorname{LN}(W_g g_v + W_\tau e_\tau).
\end{equation}
Let $\mathbf{S}_v = \{S_{v,s'}\}_{s' \in \mathcal{S}}$ denote the set of available scope representations. The scope Router compares $r_v$ with these representations and produces
\begin{equation}
\label{eq:scope_routing}
\alpha_{v,s} = p_\theta(s \mid r_v, \mathbf{S}_v).
\end{equation}
This distribution measures the relevance of each graph range to the current anchor and downstream task.

For each scope, the Router subsequently evaluates the available modality compositions. Let $\mathbf{V}_{v,s} = \{V_{v,s,m'}\}_{m' \in \mathcal{M}}$ denote the candidate set constructed within scope $s$. The scope-conditioned composition distribution is
\begin{equation}
\label{eq:composition_routing}
\beta_{v,m\mid s} = p_\theta(m \mid s, r_v, S_{v,s}, \mathbf{V}_{v,s}).
\end{equation}
Here, $S_{v,s}$ provides the shared representation of the current graph scope, whereas $\mathbf{V}_{v,s}$ contains the composition-specific candidates derived from that representation. The composition scorer combines the task-aware query and the corresponding scope representation through additive and multiplicative interactions. Consequently, the preferred modality composition can vary across graph ranges rather than being shared across all scopes.

The joint routing weight for candidate $V_{v,s,m}$ is
\begin{equation}
\label{eq:joint_routing}
\rho_{v,s,m}^{(\tau)} = \alpha_{v,s} \beta_{v,m\mid s}.
\end{equation}
This factorization couples where graph information is collected with how the modalities within that context are organized. We use $\rho_v$ to denote the resulting distribution flattened over $(s,m)$. Exact scorer definitions and a controlled comparison with the reverse composition-first factorization are provided in Supplementary Secs.~C.4 and G.2.

The routed node representation is obtained by weighted aggregation over the
candidate pool:
\begin{equation}
\label{eq:routed_mixture}
z_v^{(\tau)} = \sum_{s \in \mathcal{S}} \sum_{m \in \mathcal{M}}
\rho_{v,s,m}^{(\tau)}V_{v,s,m}.
\end{equation}

A lightweight residual Adapter then specializes the routed representation:
\begin{equation}
\label{eq:post_routing_adapter}
h_v^{(\tau)}=\operatorname{LN}\left(
z_v^{(\tau)}+\mathcal A(z_v^{(\tau)},e_\tau)\right).
\end{equation}
Both routing distributions use lightweight low-dimensional attention scorers.
Placing the Adapter after routing concentrates task-specific specialization in
a compact module without altering scope aggregation or candidate construction. The Adapter architecture and task-specific heads are detailed in Supplementary Sec.~C.5.

\definecolor{gnncolor}{RGB}{232,241,252}
\definecolor{mmgcolor}{RGB}{238,235,252}
\definecolor{gfmcolor}{RGB}{255,244,204}
\definecolor{mgfmcolor}{RGB}{255,230,230}
\definecolor{greencolor}{RGB}{232,246,235}

\newcommand{\std}[1]{{\tiny $\pm$#1}}
\newcommand{\eststd}[1]{{\tiny\textcolor{gray}{$\pm$#1}}}
\begin{table*}[t]
\centering
\renewcommand{\arraystretch}{1.10}
\setlength{\tabcolsep}{5.0pt}

\resizebox{\textwidth}{!}{
\begin{tabular}{l|cccc|cc|cc|cc}
\toprule
\multirow{3}{*}{\textbf{Method}}
& \multicolumn{4}{c|}{\textbf{NC}}
& \multicolumn{6}{c}{\textbf{LP}} \\

\cmidrule(lr){2-5}
\cmidrule(lr){6-11}

& \textbf{ele-fashion}
& \textbf{RedditS}
& \textbf{Grocery}
& \textbf{Movies}
& \multicolumn{2}{c|}{\textbf{Bili\_Food}}
& \multicolumn{2}{c|}{\textbf{DY}}
& \multicolumn{2}{c}{\textbf{sports}} \\

\cmidrule(lr){2-5}
\cmidrule(lr){6-7}
\cmidrule(lr){8-9}
\cmidrule(lr){10-11}

& \multicolumn{4}{c|}{\textbf{Acc}}
& \textbf{MRR} & \textbf{Hits@3}
& \textbf{MRR} & \textbf{Hits@3}
& \textbf{MRR} & \textbf{Hits@3} \\

\midrule
\multicolumn{11}{c}{\textit{GNN Models}} \\
\midrule

GCN
& 82.75\std{0.21}
& 91.87\std{0.28}
& 79.95\std{0.39}
& 51.13\std{0.34}
& 28.56\std{0.74}
& 35.96\std{0.98}
& 41.06\std{0.56}
& 47.79\std{0.75}
& 85.16\std{0.25}
& 96.99\std{0.12} \\

GAT
& 85.20\std{0.18}
& 91.08\std{0.25}
& 80.97\std{0.36}
& 51.22\std{0.31}
& 27.18\std{0.68}
& 33.13\std{0.89}
& 49.12\std{0.51}
& 51.56\std{0.69}
& 90.44\std{0.20}
& 96.57\std{0.09} \\

GraphSAGE
& 85.92\std{0.17}
& 92.53\std{0.22}
& 81.08\std{0.34}
& 52.31\std{0.29}
& 23.11\std{0.73}
& 23.66\std{0.96}
& 44.74\std{0.59}
& 48.59\std{0.77}
& 92.58\std{0.15}
& 96.17\std{0.07} \\

\midrule
\multicolumn{11}{c}{\textit{Multimodal Graph Models}} \\
\midrule

MMGCN
& 82.04\std{0.24}
& 81.90\std{0.31}
& 74.18\std{0.41}
& 47.56\std{0.37}
& 34.50\std{0.82}
& 42.06\std{1.05}
& 32.54\std{0.72}
& 36.50\std{0.94}
& 89.10\std{0.46}
& 91.50\std{0.24} \\

MGAT
& 85.89\std{0.15}
& 94.41\std{0.19}
& 81.93\std{0.30}
& 52.68\std{0.27}
& 37.34\std{0.61}
& 41.42\std{0.78}
& 57.14\std{0.44}
& 66.43\std{0.59}
& 90.87\std{0.22}
& 96.15\std{0.10} \\

MHGAT
& 86.05\std{0.16}
& 94.15\std{0.23}
& 81.94\std{0.32}
& \underline{53.71}\std{0.35}
& 28.18\std{0.84}
& 34.80\std{1.12}
& 50.95\std{0.63}
& 56.36\std{0.81}
& 89.29\std{0.27}
& 97.52\std{0.11} \\

DGF
& 87.04\std{0.12}
& 92.75\std{0.20}
& 82.05\std{0.28}
& 52.08\std{0.24}
& 29.77\std{0.69}
& 33.82\std{0.93}
& 54.28\std{0.47}
& 57.54\std{0.65}
& 97.61\std{0.18}
& \underline{98.91}\std{0.05} \\

\midrule
\multicolumn{11}{c}{\textit{Graph Foundation Models}} \\
\midrule

GFT (Text-Only)
& 85.66\std{0.13}
& 93.38\std{0.21}
& 81.82\std{0.26}
& 51.42\std{0.28}
& 31.38\std{0.55}
& 37.63\std{0.74}
& 62.38\std{0.61}
& 67.63\std{0.79}
& 93.34\std{0.24}
& 96.89\std{0.13} \\

RiemannGFM (Text-Only)
& 86.88\std{0.14}
& 91.41\std{0.27}
& \underline{82.01}\std{0.29}
& 51.20\std{0.31}
& 35.67\std{0.64}
& 41.22\std{0.82}
& 71.56\std{0.58}
& 76.28\std{0.76}
& 86.05\std{0.31}
& 94.45\std{0.16} \\

Graph4MM (Multimodal)
& \underline{87.06}\std{0.11}
& \textbf{95.59}\std{0.18}
& 79.48\std{0.31}
& 53.40\std{0.24}
& 38.51\std{0.52}
& 39.39\std{0.61}
& \underline{73.43}\std{0.43}
& \underline{76.50}\std{0.58}
& \underline{95.82}\std{0.21}
& 98.86\std{0.09} \\

UniGraph2 (Multimodal)
& 80.74\std{0.19}
& 93.48\std{0.22}
& 76.46\std{0.35}
& 50.70\std{0.32}
& \underline{37.66}\std{0.48}
& \underline{43.22}\std{0.57}
& 70.58\std{0.55}
& 74.29\std{0.73}
& 88.93\std{0.38}
& 93.49\std{0.19} \\

\midrule

\textbf{\textsc{BRAIN}} (Multimodal)
& \textbf{87.77}\std{0.08}
& \underline{95.34}\std{0.16}
& \textbf{82.92}\std{0.14}
& \textbf{56.25}\std{0.18}
& \textbf{38.63}\std{0.28}
& \textbf{43.64}\std{0.25}
& \textbf{75.70}\std{0.31}
& \textbf{77.76}\std{0.32}
& \textbf{97.91}\std{0.12}
& \textbf{99.09}\std{0.06} \\

\bottomrule
\end{tabular}
}

\caption{Performance comparison on node classification (NC) and link
prediction (LP). The best and second-best mean results are highlighted in
bold and underlined, respectively.}
\label{tab:baseline_comparison}
\end{table*}

\section{Experiment}
\label{sec:experiment}

We evaluate \textsc{BRAIN} through four complementary research questions: \textbf{Q1}: How effective is \textsc{BRAIN} across discriminative and generative tasks? \textbf{Q2}: How does Stage~1 pretraining affect downstream data efficiency? \textbf{Q3}: How do the Bridge, Router, and Adapter contribute to performance, and what routing behaviors do they learn? \textbf{Q4}: What computational costs does \textsc{BRAIN} incur?

Experiments cover nine multimodal graph datasets spanning node classification, link prediction, graph-to-text generation, and graph-to-image generation. Dataset statistics, task construction, leakage controls, metric definitions, and checkpoint-selection rules are documented in Supplementary Sec.~B.

\subsection{Overall Effectiveness}
\label{sec:overall_effectiveness}

\paragraph{Experimental Settings.}
We evaluate node classification (NC) on ele-fashion, RedditS, Grocery, and Movies, and link prediction (LP) on Bili\_Food, DY, and sports under the OpenMAG protocols~\cite{wan2026openmag}. We report Accuracy for NC and filtered MRR and Hits@3 for LP. For graph-to-text generation (G2Text), we use Flickr30k, where descriptions 2--5, the anchor image feature, and the permitted neighbors' text and image features are provided as input, while description 1 is held out as the generation target. We report BLEU-4, ROUGE-L, CIDEr, and BERTScore. For graph-to-image generation (G2Image), we use SemArt, where the target image is excluded from the input and used only for evaluation. We report CLIPScore and DINOv2 similarity. For fair comparison, all methods use the same task-specific generation backbones.

\paragraph{Baselines.}
For NC and LP, we compare with standard GNNs (GCN~\cite{kipf2017semi}, GAT~\cite{velickovic2018graph}, and GraphSAGE~\cite{hamilton2017inductive}), multimodal graph models (MMGCN~\cite{wei2019mmgcn}, MGAT~\cite{tao2020mgat}, MHGAT~\cite{jia2023mhgat}, and DGF~\cite{zheng2026dgf}), graph foundation models (GFT~\cite{wang2024gft} and RiemannGFM~\cite{sun2025riemanngfm}), and multimodal graph foundation models (UniGraph2~\cite{he2025unigraph2}, Graph4MM~\cite{ning2025graph4mm}, and PLANET~\cite{liu2026planet}). For generation, NTSFormer~\cite{hu2026ntsformer}, PLANET, and Graph4MM are evaluated where applicable within the same task-specific language or diffusion pipeline, with matched data splits, context budgets, prompts, and decoding or sampling settings.

\begin{figure}[!t]
    \centering
    \includegraphics[width=0.90\columnwidth]{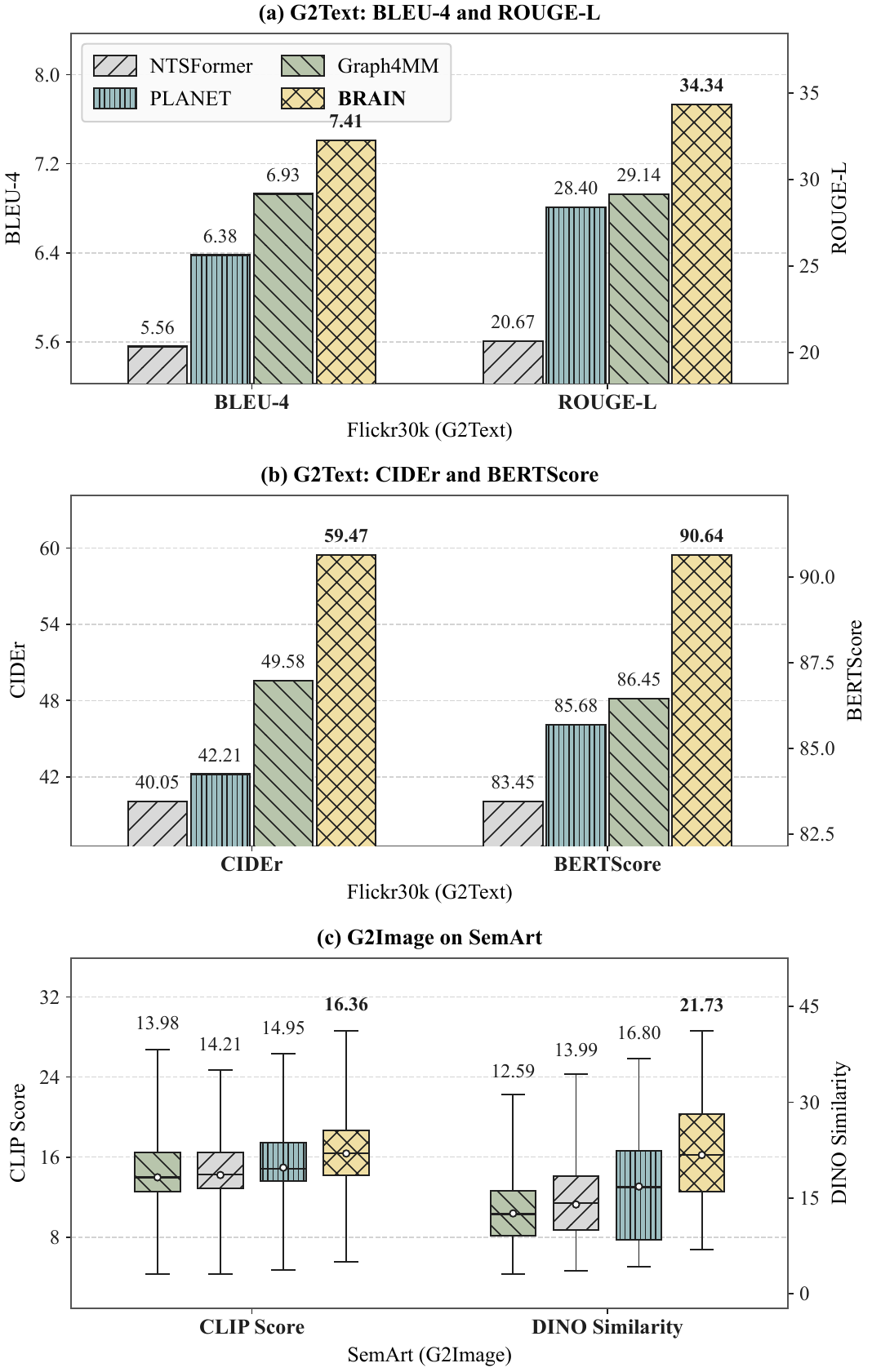}
    \vspace{1.0mm}
    \caption{Generative performance on Flickr30k G2Text and SemArt G2Image.
    We report BLEU-4, ROUGE-L, CIDEr, and BERTScore for G2Text, and
    CLIPScore and DINOv2 similarity for G2Image. Relative improvements
    are computed against the strongest baseline for each metric.}
    \label{fig:main_generation}
\end{figure}

\paragraph{Discriminative tasks.}
Tab.~\ref{tab:baseline_comparison} shows that BRAIN ranks first on nine of the ten dataset--metric comparisons and second on the remaining one. Its strongest NC advantage appears on Movies, while it remains competitive on RedditS and Grocery. BRAIN also leads on every reported LP dataset and metric.These results are consistent with the value of jointly modeling scope and
modality composition; the causal contribution of this design is examined through controlled ablations below. The smaller margins on some NC datasets further suggest that the largest gains occur when a fixed local view is insufficient, rather than being tied to a single favorable domain.

\paragraph{Generative tasks.}
As shown in Fig.~\ref{fig:main_generation}, BRAIN leads all baselines across G2Text and G2Image metrics, outperforming the top baseline by 19.9\% in CIDEr and 29.3\% in DINO Similarity. Gains in G2Text reflect better lexical coverage and semantic fidelity, while higher CLIPScore and DINOv2 demonstrate superior text--image alignment and visual consistency.(Detailed analysis are given in Supplementary Sects.~E.2 and I.)

\subsection{Data Efficiency of Stage-1 Pretraining}
\label{sec:transferability}

We compare random initialization with joint multi-graph Stage~1 pretraining while fixing the Stage~2 architecture, optimization schedule, and data split. Because the target graphs participate in Stage~1, this experiment measures downstream data efficiency rather than transfer to an unseen graph. We vary Stage~2 supervision from 5\% to 80\% on Movies NC and Bili\_Food LP, and from 10\% to 80\% on DY LP; exact subsampling protocols and complete values are reported in Supplementary Sec.~F.

\begin{figure}[t]
    \centering
    \includegraphics[width=1.0\linewidth]{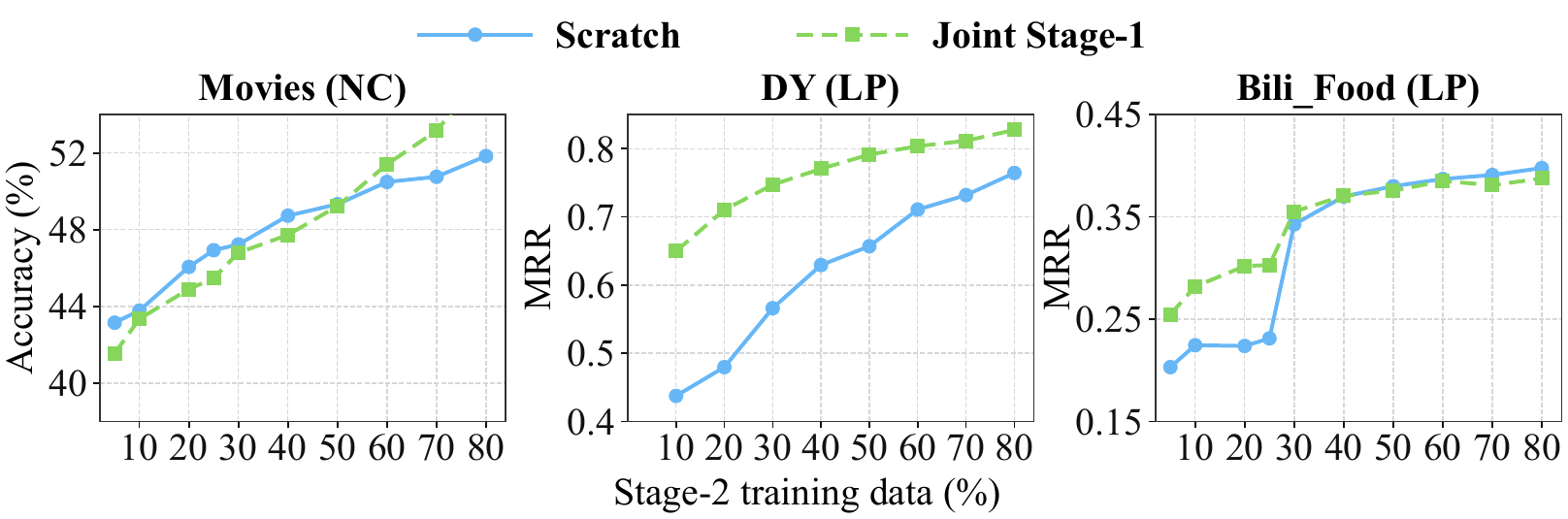}
    \caption{Data efficiency on Movies NC (left), DY LP (middle), and Bili\_Food LP (right). Random initialization and joint Stage~1 pretraining use the same Stage~2 protocol.}
    \label{fig:low_resource}
\end{figure}

\paragraph{Reduced-supervision results.}
Figure~\ref{fig:low_resource} shows that Stage~1 most consistently improves data efficiency on LP. On DY, pretrained initialization improves MRR at every supervision level, including gains of 0.2123 and 0.2308 at 10\% and 20\% supervision. Bili\_Food obtains its largest gain, 0.0781 MRR, at 20\%. These results indicate that multi-graph pretraining provides a useful relational prior, although its benefit depends on the downstream task and the amount of supervision.

Movies exhibits a different adaptation profile: pretrained initialization is slightly weaker in the lowest-supervision regimes, but the gap narrows and reverses as more labels become available. At 80\% supervision, Accuracy rises from 51.84\% to 53.96\%, an absolute gain of 2.12 percentage points. This crossover suggests that transferable scope representations may require sufficient task supervision to be specialized effectively, rather than providing a uniform advantage in every extreme low-label regime.

\subsection{Ablation and Routing Analysis}
\label{sec:mechanism}

We analyze the necessity of each module using component ablations, and investigate the learned scope--composition distributions across representative datasets.

\paragraph{Experimental settings.}
The Bridge controls remove graph context while retaining the anchor representation, collapse the three modality-composition candidates within each scope, or retain only the 1-hop scope. The routing controls replace the conditional factorization $\alpha_s\beta_{m\mid s}$ with three alternatives: \emph{uniform mixing} assigns weight $1/9$ to every candidate; the \emph{independent Router} uses $\alpha_s\beta_m$ and shares one composition distribution across all scopes; and the \emph{flat 9-way Router} treats the nine scope--composition candidates as unrelated choices by predicting one softmax directly over them. The final control removes the post-routing residual Adapter. Supplementary Sec.~G additionally examines routing distributions, factorization order, context budgets, and sparse route selection.

\begin{figure*}[!t]
    \centering
    \includegraphics[width=0.98\textwidth]{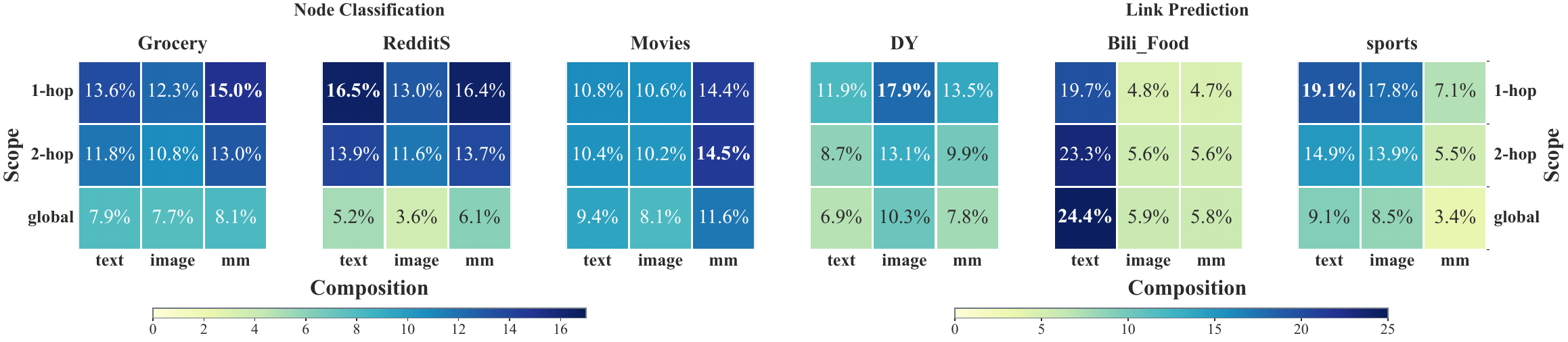}
    \caption{Average scope--composition routing distributions across representative NC and LP datasets. The preferred graph range and modality composition vary across datasets and tasks.}
    \label{fig:router_preference}
\end{figure*}

\begin{table}[!ht]
    \centering
    \setlength{\tabcolsep}{4.0pt}
    \renewcommand{\arraystretch}{1.10}
    \resizebox{\columnwidth}{!}{%
    \begin{tabular}{lccccc}
        \toprule
        \multirow{2}{*}{Method}
        & \multicolumn{3}{c}{Node Classification}
        & \multicolumn{2}{c}{Link Prediction} \\
        \cmidrule(lr){2-4}
        \cmidrule(lr){5-6}
        & Grocery & RedditS & Movies & Bili\_Food & DY \\
        \midrule
        \multicolumn{6}{c}{\textit{Bridge and candidate construction}} \\
        w/o graph context
        & 79.31 & 91.92 & 49.31 & 31.78 & 61.40 \\
        w/o composition modulation
        & 82.45 & 94.31 & 52.22 & 34.16 & 59.23 \\
        Single scope (1-hop only)
        & 82.16 & 93.91 & 52.97 & 33.85 & 62.36 \\
        \midrule
        \multicolumn{6}{c}{\textit{Routing mechanism}} \\
        Uniform mixing $(1/9)$
        & 81.69 & 94.23 & 52.67 & 35.17 & 70.50 \\
        Independent Router $(\alpha_s\beta_m)$
        & 82.60 & 94.31 & 53.46 & 35.89 & 74.16 \\
        Flat 9-way joint Router
        & 82.78 & 94.22 & 54.43 & 36.58 & 74.37 \\
        \midrule
        \multicolumn{6}{c}{\textit{Adaptation mechanism}} \\
        w/o Adapter
        & 82.60 & 94.47 & 52.79 & 29.46 & 61.57 \\
        \midrule
        \textbf{\textsc{BRAIN} (full)}
        & \textbf{82.92} & \textbf{95.34} & \textbf{56.25}
        & \textbf{38.63} & \textbf{75.70} \\
        \bottomrule
    \end{tabular}%
    }
    \caption{Ablation results on representative NC and LP datasets. NC reports Accuracy (\%); LP reports MRR (\%).}
    \label{tab:Brain_ablation}
\end{table}

\paragraph{Ablation findings.}
Removing graph context causes the largest and most consistent degradation, confirming that neighborhood information contributes beyond the anchor modalities. Collapsing composition candidates or retaining only one scope is particularly damaging on LP and Movies, where useful signals are distributed across graph ranges and modality configurations. All three routing controls remain below the Hierarchical Context Router in the reported results, while removing the Adapter produces a pronounced LP drop. These results support these modules as necessary components: the Bridge constructs complementary candidates, the Router performs context-dependent selection, and the Adapter specializes the routed representation.
Together, these results indicate that BRAIN's gains do not arise merely from enlarging the candidate bank: scope-preserving construction, conditional routing, and post-routing specialization contribute complementary benefits across classification and relational prediction.

\paragraph{Routing preferences.}
Fig.~\ref{fig:router_preference} reveals dataset-specific routing patterns. Grocery and RedditS favor local context, whereas Movies balances different scopes while preferring multimodal candidates. DY is image-dominant, Bili\_Food favors broader text-dominant context, and Sports relies mainly on local text and image evidence. These patterns show that neither a fixed graph range nor a single modality composition is optimal across datasets.

\subsection{Efficiency and Complexity}
\label{sec}

\paragraph{Complexity.}
\textsc{BRAIN} contains 7.33M trainable parameters in the profiled NC configuration, of which the Router accounts for only 8.3\%. With three scopes and three modality compositions, the candidate bank has a fixed size of $K=9$, and Router scoring grows linearly with the number of scope--composition pairs. The main additional cost arises from constructing and scoring the nine candidates; the lightweight Adapter is evaluated only once after routing. Therefore, the candidate bank is the principal target for future top-$k$ or hard-routing acceleration. Formal complexity and profiling boundaries are detailed in Supplementary Sec.~H.


\begin{figure}[t]
    \centering
    \includegraphics[width=1.0\linewidth]{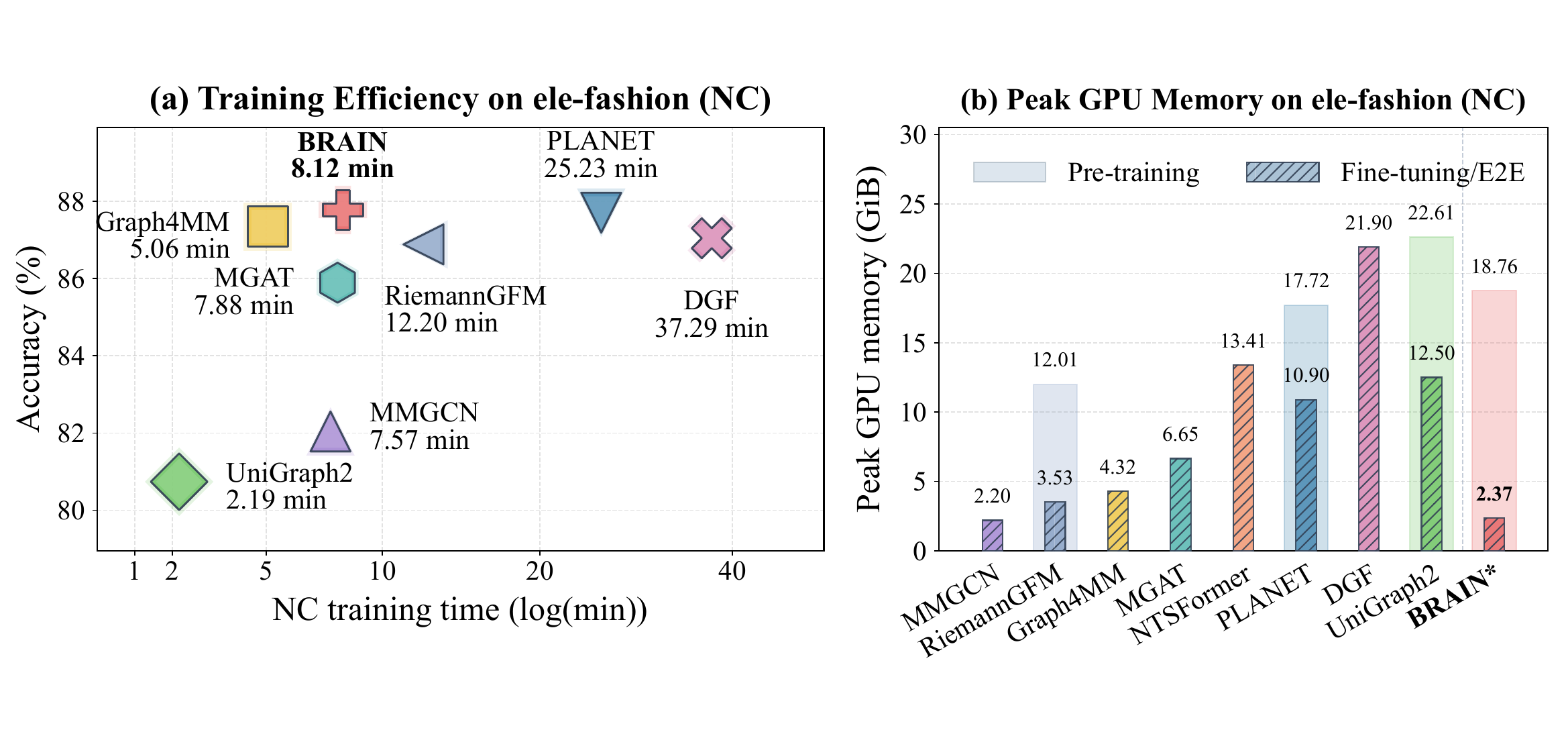}
    \caption{Efficiency and memory consumption analysis.}
    \label{fig:efficiency_memory}
\end{figure}

\paragraph{Efficiency and memory profiling.}
As shown in Fig.~\ref{fig:efficiency_memory}, \textsc{BRAIN} achieves the highest accuracy with a Stage~2 training time of 8.12 minutes, demonstrating a favorable accuracy--efficiency trade-off. Although several lightweight baselines train faster, they obtain lower accuracy.

\textsc{BRAIN} also requires only 2.37 GiB of peak GPU memory during Stage~2, the second-lowest footprint among all compared methods and only 0.17 GiB above MMGCN. This is 32.9\%--89.2\% lower than the remaining baselines. Although Stage~1 pretraining peaks at 18.76 GiB, Stage~2 reduces the memory requirement by 87.4\%. These results show that learning multiple scope--composition candidates is a manageable one-time pretraining cost while preserving efficient downstream adaptation.

\section{Conclusion}
\label{sec:conclusion}

We introduced \textsc{BRAIN}, a multimodal graph foundation model for graph-conditioned context relevance. \textsc{BRAIN} combines attention-based semantic aggregation and GCN-based structural propagation to construct a candidate pool of graph scopes and modality compositions. A hierarchical conditional Router mixes these candidates, after which a residual Adapter specializes the routed representation for downstream prediction. Its two-stage training first learns reusable multimodal and structural representations across graphs and then optimizes candidate construction, routing, adaptation, and task heads with supervision. Across nine datasets and four task families, \textsc{BRAIN} achieves strong performance, with clear gains on relational matching and generation. Low-resource experiments demonstrate task-dependent data-efficiency benefits, while ablations and routing analyses validate graph context, structured candidate construction, conditional selection, and domain-specific context preferences across heterogeneous datasets and downstream tasks.

\bibliography{References}

\end{document}